\documentclass[runningheads]{llncs}
\usepackage{graphicx}
\usepackage{cite}
\usepackage{subfigure}
\usepackage{url}
\usepackage{verbatim,epsf}
\usepackage[font=footnotesize]{subfig}

\usepackage{multicol}
\usepackage{multirow}
\usepackage[font=footnotesize,labelfont=bf]{
	caption}
\def\tablename{Table}
\usepackage{siunitx}
\usepackage{bm}
\usepackage{subfig}
\usepackage{graphicx}

\usepackage{tabularx}
\usepackage{balance} 

\usepackage{amssymb}
\usepackage{amsmath}
\usepackage{graphicx}
\usepackage{epstopdf}

\usepackage{epsfig, color}

\usepackage{comment} 
\usepackage{amsfonts} 
\usepackage{amsmath} 
\usepackage{rotating}

\usepackage{amssymb}
\usepackage{amsmath}
\usepackage{amsfonts}
\usepackage{wasysym}

\usepackage{epsfig, color}

\begin{document}
\title{R2P: A Deep Learning Model from mmWave Radar to Point Cloud}
%
%

\author{Yue Sun\inst{1}\orcidID{0000-0002-4284-6309} \and
Honggang Zhang\inst{2}\orcidID{0000-0002-5311-6520} \and
Zhuoming Huang\inst{3}\orcidID{0000-0002-5105-6501} \and
Benyuan Liu\inst{4}\orcidID{0000-0001-5879-4702}}
\authorrunning{Y. Sun}
%
\institute{The Department of Computer Science, UMass Boston, USA\\
	\email{Yue.Sun001@umb.edu}\and
	The Department of Engineering, UMass Boston, USA\\
	\email{Honggang.Zhang@umb.edu}\and
	The Department of Engineering, UMass Boston, USA\\
	\email{Zhuoming.Huang001@umb.edu}\and
	The Department of Computer Science, UMass Lowell, USA\\
	\email{bliu@cs.uml.edu}
}
\maketitle              
%
\begin{abstract}
Recent research has shown the effectiveness of mmWave radar sensing for object detection
in low visibility environments,
which makes it an ideal technique in autonomous navigation systems.
In this paper, we introduce Radar to Point Cloud (R2P), 
a deep learning model that generates smooth, dense, and highly accurate point cloud representation of 
a 3D object with fine geometry details, 
based on rough and sparse point clouds with incorrect points obtained from mmWave radar. 
These input point clouds are  
converted from the 2D depth images that are generated from raw mmWave radar sensor data, characterized by  
inconsistency, and orientation and shape errors.
R2P utilizes an architecture of two sequential deep learning encoder-decoder blocks 
to extract the essential features of those radar-based input point clouds of an object when observed from
multiple viewpoints, and to ensure the internal consistency of a generated output point cloud and 
its accurate and detailed shape reconstruction of the original object.   
We implement R2P to replace the Stage 2 of our recently proposed 3DRIMR (3D Reconstruction and Imaging via 
mmWave Radar) system.
Our experiments demonstrate the significant performance improvement of R2P over 
the popular existing methods such as PointNet,
PCN,
and the original 3DRIMR design.




{}

\end{abstract}
%
%

\section{INTRODUCTION}

Recently the advantage of Millimeter Wave (mmWave) radar in object sensing in low visibility environment has 
been actively studied and applied in autonomous 
vehicles \cite{HawkEye} and search/rescue in high risk areas \cite{mobisys20smoke}.
However, further application of mmWave radar in object imaging and reconstruction is quite difficult because of 
the characteristics of mmWave radar signals such as low resolution, sparsity, and large noise due to multi-path and 
specularity. Recent work \cite{HawkEye, superrf,mobisys20smoke} attempts to design deep learning systems 
to generate 2D depth images based on mmWave radar signals. 3DRIMR \cite{sun20213drimr} further introduces 
an architecture that generates 3D object shapes based on mmWave radar, but there is still room for significant improvement 
in order to produce  more satisfactory end results. 

In this paper, we introduce Radar to Point Clouds (R2P), a deep learning model 
that generates 3D objects in the format of dense and smooth point clouds that accurately resemble the 3D shapes of original objects
with fine details.
The inputs to R2P are simply some rough and sparse point clouds with possibly many inaccurate points 
due to the imperfect conversion from raw mmWave radar data. 
For example, they can be 
directly converted from those 2D depth images generated from radar data, using systems such as \cite{HawkEye} 
or the Stage 1 of 3DRIMR \cite{sun20213drimr}.
R2P can be used to replace the generator network in Stage 2 of 3DRIMR.
Recall that 3DRIMR's Stage 1 
takes 3D radar intensity data as input and generates 2D depth images of an object,
which are then combined and converted to a rough and sparse point cloud to be used as 
the input to Stage 2. The generator network of Stage 2  
outputs the 3D shape of the object in the form of dense and smooth point cloud.  
Even though 3DRIMR can give some promising results, its Stage 2's generator network design is still not quite 
satisfactory. Specifically, the edges of generated point clouds are still blurry and their points 
tend to be evenly distributed in space and thus do not give a clear sharp shape structure. 
We introduce R2P in this paper to replace the Stage 2 of the original 3DRIMR, and we find that R2P 
significantly outperforms 3DRIMR both quantitatively and visually. 
	Our major contributions are summarized as follows.
		We propose Radar to Point Cloud (R2P),  a deep neural network model, 
		to generate smooth, dense, and highly accurate point cloud representation of 
		a 3D object with fine geometry details, 
		based on rough and sparse point clouds with incorrect points. These input point clouds are directly 
		converted from the 2D depth images of an object that are generated from raw mmWave radar sensor data, and thus characterized by  
		mutual inconsistency and orientation/shape errors, due to the imperfect process to generate them.	
			R2P utilizes an architecture of two sequential deep learning encoder-decoder blocks 
			to extract the essential features of those input point clouds of an object when observed from
			various viewpoints, and to ensure the internal consistency of a generated output point cloud and 
			its detailed shape reconstruction of the original object.   
			We further demonstrate with extensive experiments the importance of loss function design 
			in the training of models for reconstructing point clouds. We also show
			the limitations of Chamfer Distance (CD) and Earth Mover's Distance (EMD),			
			 the two state-of-the-art point clouds evaluation metrics, in the evaluation of the shape similarity of two point clouds.

In the rest of the paper, we discuss related work and preliminaries in Sections 
	\ref{sec_related} and \ref{sec_background}, and then the design of R2P in Section \ref{sec_design}. 
	Experiment results are given in Section \ref{sec_imp}. 
	Finally the paper concludes in Section \ref{sec_conclusion}.

\section{RELATED WORK}\label{sec_related}
Frequency Modulated Continuous Wave (FMCW) Millimeter Wave (mmWave) radar sensing 
has been an active research area in recent years, especially in applications such as person/gesture identification \cite{vandersmissen2018indoor,yang2020mu}, car detection/imaging \cite{HawkEye}, and environment sensing \cite{mobisys20smoke, superrf}. Usually Synthetic Aperture Radar (SAR) is used in data collection 
for high resolution, e.g., \cite{mamandipoor201460,national2018airport,ghasr2016wideband,sheen2007near}.
This paper is built on our recent work \cite{sun20213drimr} 
on applying mmWave radar for 3D object reconstruction, in which we proposed 3DRIMR. 
The deep neural network model proposed in this paper can replace the model in the Stage 2 of 3DRIMR, 
and this new model significantly outperforms the original 3DRIMR. 
There have been a few recent work on mmWave radar based imaging, mapping, and 3D object 
reconstruction \cite{HawkEye,superrf,mobisys20smoke,yuan2018pcn,qi2016pointnet}. Our work is inspired by 
their promising research results, especially PointNet \cite{qi2016pointnet,qi2017pointnet++}
and PCN \cite{yuan2018pcn}. Due to the low cost and small form factor of commodity mmWave 
radar sensors, we plan to develop a simple 3D reconstruction system with fast data collection to 
be attached in our UAV SLAM system \cite{sun2020lidaus} for search and rescue in dangerous environment.
Besides radar signals, vision community has also been working on learning-based 
3D object shape reconstruction \cite{yang20173d,dai2017shape,sharma2016vconv,smith2017improved}, 
most of which use voxels to represent 3D objects.
Our proposed neural network model  
uses point cloud as a format for 3D objects to capture detailed geometric information 
with efficient memory and computation performance. 
Our proposed R2P significantly outperforms the existing methods such as PointNet \cite{qi2016pointnet}, 
PointNet++ \cite{qi2017pointnet++}, PCN\cite{yuan2018pcn}, and 3DRIMR \cite{sun20213drimr}.

\section{PRELIMINARIES}\label{sec_background}

\subsection{FMCW Millimeter Wave Radar Sensing and Imaging}

Similar to \cite{sun20213drimr}, we use FMCW mmWave radar 
sensor \cite{timmwave} signals to reconstruct 3D object shapes. 
Fast Fourier Transform (FFT) along three directions are conducted  on received waveforms
to generate 3D intensity maps of a space that
represent the energy or radar intensity per voxel in the space, written as $x(\phi, \theta, \rho)$.
Note that  $\phi$, $\theta$, and $\rho$ represent azimuth angle, elevation angle, and range respectively. 
We use IWR6843ISK \cite{iwr6843} operating at $60$ GHz frequency,
and for high resolution radar signals, we adopt SAR operation.
Unlike data from LiDAR and camera sensor, mmWave radar sensors can only give us 
sparse, low resolution, and highly noisy data. Incorrect ghost points in radar signals 
can be generated due to multi-path effect. References \cite{HawkEye,superrf,mobisys20smoke} give more detailed discussion 
on FMCW mmWave radar sensing. 

\subsection{Representation of 3D Objects}
We adopt point cloud format to represent 3D objects. 
Even though point cloud is a standard representation of 3D objects and it is used in 
learning-based 3D reconstruction, e.g., \cite{fan2017point,qi2016pointnet,qi2017pointnet++}, the convolutional operation
of Convolutional Neural Network (CNN) cannot be directly applied to a point cloud set as it is 
essentially an unordered point set.  Furthermore, the point cloud of an object that is directly generated by 
raw radar signals is not a good choice to represent the shape of an object because of radar signal's 
low resolution, sparsity, and incorrect ghost points due to multi-path effect. 
Besides point clouds,  voxel-based representations can also be used in 3D reconstruction \cite{kar2017learning,paschalidou2018raynet,ji2017surfacenet,wu2016learning}, 
and the advantage of such representation is that 3D CNN convolutional operations can be applied.
In addition, mesh representations of 3D objects are also used in existing work \cite{kong2017using,wang2018pixel2mesh}.
However these two representation formats are limited by memory and computation cost.

\subsection{Review of 3DRIMR Architecture}\label{review_3drimr}
This paper introduces R2P as a generator network to replace the Stage 2 of 3DRIMR
in order to generate smooth and dense point clouds.
For completeness, we now briefly review 3DRIMR's architecture. 
3DRIMR consists of two back-to-back generator networks 
$\mathbf{G_{r2i}}$ and $\mathbf{G_{p2p}}$.
In Stage 1, $\mathbf{G_{r2i}}$ receives a 3D radar energy intensity map
of an object and outputs a 2D depth image of the object.
We let a mmWave radar sensor scans an object from multiple viewpoints
to get multiple 3D energy maps. Then  $\mathbf{G_{r2i}}$ generates 
the corresponding 2D depth
images of the object. 
The Stage 2 of 3DRIMR first pre-processes these images to get multiple coarse point clouds of the object,
which are used as the input to $\mathbf{G_{p2p}}$ to generate a single point cloud of 
the object. 
A conditional generative adversarial network (GAN) architecture is designed for 3DRIMR's training. That is,
two discriminator networks $\mathbf{D_{r2i}}$ and $\mathbf{D_{p2p}}$ that are jointly trained together with their corresponding generator networks.
Let $m_r$ denotes a 3D radar intensity map of an object captured 
from a viewpoint, and 
let $g_{2d}$ be a ground truth 2D depth image of the same object captured 
from the same viewpoint. 
$\mathbf{G_{r2i}}$ generates $\hat{g}_{2d}$ that predicts or estimates $g_{2d}$ given $m_{r}$. 
If there are $k$ different viewpoints $v_1, ..., v_k$, 
generator $\mathbf{G_{r2i}}$ predicts their corresponding 2D depth images $\{\hat{g}_{2d,i} | i = 1, ..., k\}$.
Each $\hat{g}_{2d,i}$ can be directly converted to a coarse and sparse 3D point cloud. 
Then we can have $k$ coarse point clouds $\{P_{r,i} | i = 1, ..., k\}$
of the object. 
The Stage 2 of 3DRIMR unions the $k$ coarse point clouds 
to form an initial estimated coarse point cloud of the object, denoted as $P_r$, which is 
a set of 3D points $\{p_j | j = 1, ..., n\}$.
Generator $\mathbf{G_{p2p}}$ takes 
$P_r$ as input, and outputs a dense and smooth point cloud $P_o$.
Note that since the generation process of $\mathbf{G_{r2i}}$ is not perfect, 
a coarse $P_r$ may likely contain many missing or even incorrect points. 
Next we will discuss our proposed R2P.

\section{R2P Design}\label{sec_design}

\subsection{Overview}

\begin{figure}
	\includegraphics[width=\textwidth]{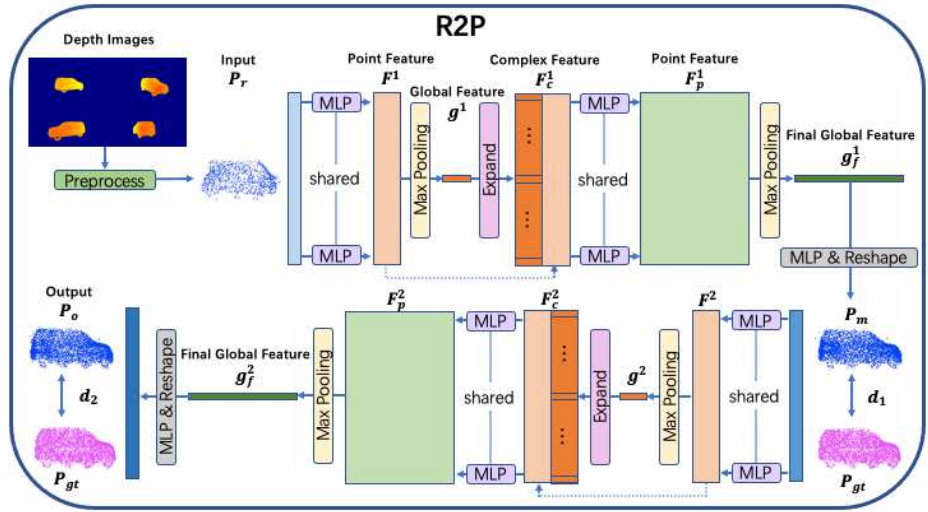}
	\caption{R2P network architecture. 
		The system first pre-processes multiple 2D depth images of an object   
		to get multiple coarse and sparse point clouds, which may contain many incorrect points. 
		These depth images are generated for the same object but viewed from four different viewpoints.
		Combining those coarse point clouds we can derive a single coarse point cloud, 
		which is used as the input of R2P. Then 
		R2P outputs a dense, smooth, and accurate point cloud representation of the object.} \label{fig_G}
\end{figure}

R2P is a generative model that generates a smooth and dense 3D point cloud of an object 
from the union of multiple coarse point clouds, which are directly 
converted from the 2D depth images of the object observed from different viewpoints.
Since those 2D depth images are generated from raw radar energy maps, the union of
their converted point clouds may contain many inconsistent and incorrect points due to the imperfect image generation process. 
R2P network model is able to extract the essential features of those input point clouds, 
and to ensure the internal consistency of a generated output point cloud 
and its detailed, accurate shape reconstruction of the original object.
R2P can be used to replace the generator network of the Stage 2 of 3DRIMR, 
and it can also be used as an independent network that works on any rough and sparse input point clouds that contain incorrect points. 
The architecture of R2P is shown in Fig. \ref{fig_G}. 
Let $P_r$ denote a union of $k$ separate coarse point clouds observed from $k$ viewpoints of the object $\{P_{r,i} | i=1,...,k\}$.
R2P aims at generating a point cloud of an object 
with continuous and smooth contour $P_o$ from $P_r$.
In our design we also include a discriminator network to train the generator under GAN framework. Due to space limitations, 
we do not discuss the discriminator here. 


\subsection{R2P Architecture}

R2P's input $P_r$ and output $P_o$
are 3D point clouds represented as $n \times 3$ matrices, where $n$ is the number of points in the point cloud, 
and each row represents the 3D Cartesian coordinate $(x, y, z)$ of a point. 
Note that the output point cloud generated from our network has a larger number of points than the input point cloud, 
i.e., our network can reconstruct a dense and smooth point cloud from a sparse one.
R2P consists of two sequential processing blocks, and both blocks share the same encoder-decoder network design.
The first block takes the raw input point cloud $P_r$ to produce 
an intermediate point cloud $P_m$, and then the second block processes $P_m$ to generate the final output $P_o$.
In each block, 
the encoder takes its input point cloud and converts it to a high dimensional feature vector, 
and the decoder takes this feature vector and converts it to an intermediate or a final output point cloud.
\noindent \textbf{Encoder.}
As shown in Fig. \ref{fig_G}, in the first block, the encoder gets $P_r$  as input, 
and passes it to a shared multilayer perceptron (MLP) to extend every point $p_i$ in $P_r$ to a high dimensional feature vector $f_i$ 
and form the point feature matrix $F^1$. 
This shared MLP is a network with two linear layers with BatchNorm and ReLU in between.
Then, the encoder applies a point-wise maxpooling on $F^1$ and extracts a global feature vector $g^1$.
To produce a complete point cloud for an object, 
we need both local and global features, 
therefore, the encoder concatenates the global feature with each of the point features
$f^1_i$ (of $F^1$) and form the complex feature matrix $F^1_c$. Then another shared MLP is used to produce 
point feature matrix $F^1_p$.
Then, another point-wise maxpooling will perform on $F^1_p$ to extract the final global feature $g^1_f$. 

\noindent \textbf{Decoder.}
Decoder takes the final global feature $g^1_f$ as input,
and then passes it to a MLP, which consists of three fully-connected layers with ReLU in between.
After this MLP, the final global feature is converted to $1 \times 3m$ vector (where $m$ is the number of points in $P_m$), 
and then reshaped to $m \times 3$ matrix which represents the point cloud $P_m$. 

As shown in Fig. \ref{fig_G}, the second block's design is similar to the first block.  The encoder of the second block
takes $P_m$ as input to produce a final global feature $g^2_f$, and then the decoder generates 
the final output point cloud $P_o$.

\subsection{Loss Function}

Due to the irregularity of point clouds, 
it is quite difficult to choose an effective loss function to indicate the difference between a generated point cloud and 
its corresponding ground truth point cloud.
There are two popular metrics to evaluate the difference between two point clouds:
Chamfer Distance (CD) \cite{yuan2018pcn} and Earth Mover's Distance (EMD) \cite{yuan2018pcn}.
CD calculates the average closest distance between two point clouds $S_1$ and $S_2$.
The symmetric version of CD is defined as:
\begin{equation}
\small{
	CD(S_1, S_2) = 
	\frac{1}{|S_1|} \sum_{x \in S_1} \mathop{min}\limits_{y \in S_2} \lVert x-y \rVert_{2} 
	+ \frac{1}{|S_2|} \sum_{y \in S_2} \mathop{min}\limits_{x \in S_1} \lVert y-x \rVert_{2}
}
\end{equation}
EMD can find a bijection $\phi: S_1 \to S_2$, 
which can minimize the average distance between each pair of corresponding points in two point clouds $S_1$ and $S_2$. 
EMD is calculated as:
\begin{equation}
\small{
	EMD(S_1, S_2) =  
	\mathop{min}\limits_{\phi: S_1 \to S_2} \frac{1}{|S_1|}
	\sum_{x \in S_1} || x-\phi(x) ||_{2} 
}
\end{equation}
In our system, R2P generates the intermediate point cloud $P_m$ after its first encoder-decoder block, 
and then generates the final output point cloud $P_o$ after the second block.
Hence, we design our loss function to evaluate both generated point clouds. That is, 
R2P's loss function is defined as a weighted sum of the loss of the first block $d_1(P_m, P_{gt})$ 
and the loss of the second block $d_2(P_o, P_{gt})$. 
\begin{equation}
	L(R2P) =  d_1(P_m, P_{gt}) + \alpha d_2(P_o, P_{gt}) 
\end{equation}
Both $d_1$ and $d_2$ can be either CD or EMD, or some weighted combination of them. 
$\alpha$ is a hand-tuned parameter, and in our experiments, we find it performs well when set to 0.1.
Note that unlike CD, EMD needs to find a bijection relationship between two point clouds, 
which is a optimization problem and hence computationally expensive, especially when the point clouds have large amounts of points.
Moreover, this bijection also requires these two evaluated point clouds have the same number of points.

\section{IMPLEMENTATION AND EXPERIMENTS}\label{sec_imp}

We implement our proposed R2P network and use it as the generator network of  
3DRIMR system's Stage 2. 
The system first generates 2D depth images from 3D radar intensity maps from multiple views of an object,
and then passes these output depth images to R2P to produce a 3D point cloud of the object.

\subsection{Datasets}
%
We conduct experiments on a dataset including 5 different categories of objects, 
namely industrial robot arms, cars, chairs, desks, and L-shape boxes.
This dataset consists of both synthesized data and real data collected from experiments.
The input point clouds to R2P are the output depth images produced by 3DRIMR's Stage 1.
We follow a procedure that is similar to \cite{sun20213drimr} to generate ground truth point clouds.
%
In order to obtain enough amount of data required to train our deep neural network models,
we modified HawkEye's data synthesizer \cite{web:hawkeyeCode} to synthesize 3D radar intensity maps and 2D depth images from 3D CAD point-cloud models, with configurations matching TI's mmWave radar sensor IWR6843ISK with DCA1000EVM \cite{dca1000evm}, 
and Stereolabs' ZED mini camera \cite{zed}.
In our experiments and when generating synthesized data, 
we determine the space setup to capture the views of various objects in a 3D environment. 
Four pairs of mmWave radar sensors and depth camera sensors are placed at the centers of the four edges of a square area, 
pointing towards the center of the square area where objects are randomly placed. 

\subsection{Model Training and Testing}
We use the 2D depth images generated in the Stage 1 of 3DRIMR 
to form a dataset of coarse and sparse input point clouds for the $5$ different categories of objects.
Each input point cloud has $1024$ points and an intermediate/final output point cloud has $4096$ points.
We train our proposed network model for each category independently.
For each object category, we train the model based on $1400$ pairs of point clouds for $200$ epochs with batch size $2$.
The learning rate for the first 100 epochs is $2 \times 10^{-4}$ and linearly decreases to 0 in the rest 100 epochs.
Then we test the model using the remaining $100$ point clouds.


\subsection{Evaluation Results}

To the best of our knowledge, except for our previous work 3DRIMR, 
there are no other point cloud-based networks to 
reconstruct smooth, dense, and accurate point clouds from \textit{point clouds with many incorrect and inconsistent points}, 
e.g., the union of multiple coarse point clouds which are converted from various 2D depth images of an object 
that are possibly inaccurate and inconsistent between themselves in terms of orientation and shape structure details.
For example, a generated 2D depth image from radar data \cite{HawkEye,sun20213drimr} can possibly
have a car's left and right sides switched, or it can be shaped like a similar but different car with different shape details. 
Note that the existing works (e.g., \cite{yuan2018pcn}) on this subject usually do not have such strong inaccuracy assumption on their inputs except missing points.   
Nevertheless we compare our proposed architecture against five related baseline methods. 
They are popular point cloud-based generative models, mainly designed for the purposes 
of classification, segmentation, and point clouds completion, assuming input data is either sparse or there are missing points.  
To ensure fair comparison, we train all six models (including baselines and ours) 
based on the same training and testing datasets.
These baseline methods include: 
	(1) \textbf{PointNet}.
	It is introduced in \cite{qi2016pointnet}. We choose $1024$ as the dimension of global feature of PointNet. 
	The loss function is a combination of CD and EMD.
	(2) \textbf{PointNet++}.
	We use the same encoder architecture of PointNet++ \cite{qi2017pointnet++} for classification with three Set Abstraction (SA) modules to get a 1024-dimension global feature, followed by a decoder consists of 3 fully-connected layers.  
	The loss function is also a combination of CD and EMD.
 	(3) \textbf{PCN\_CD}.
	We use the same architecture of PCN\cite{yuan2018pcn}, and set the grid size as 2. 
	The number of points in the coarse and detailed output point clouds are 1024 and 4096 respectively. 
	Note that in this method, both $d_1$ and $d_2$ are CD.
	(4) \textbf{PCN\_EMD}.
	In this method, the architecture of the network is same as PCN\_CD, 
	but $d_1$ is EMD.
	(5) \textbf{3DRIMR}.
	This is the only architecture designed for point clouds reconstruction from inaccurate input point clouds among the $5$ baselines.
	To ensure fair comparison, we use the combination of CD and EMD as its loss function.
\begin{figure*}[htb!]
	\centerline{
		\begin{minipage}{1\linewidth}
			\begin{center}
				\setlength{\epsfxsize}{0.95\linewidth}
				\epsffile{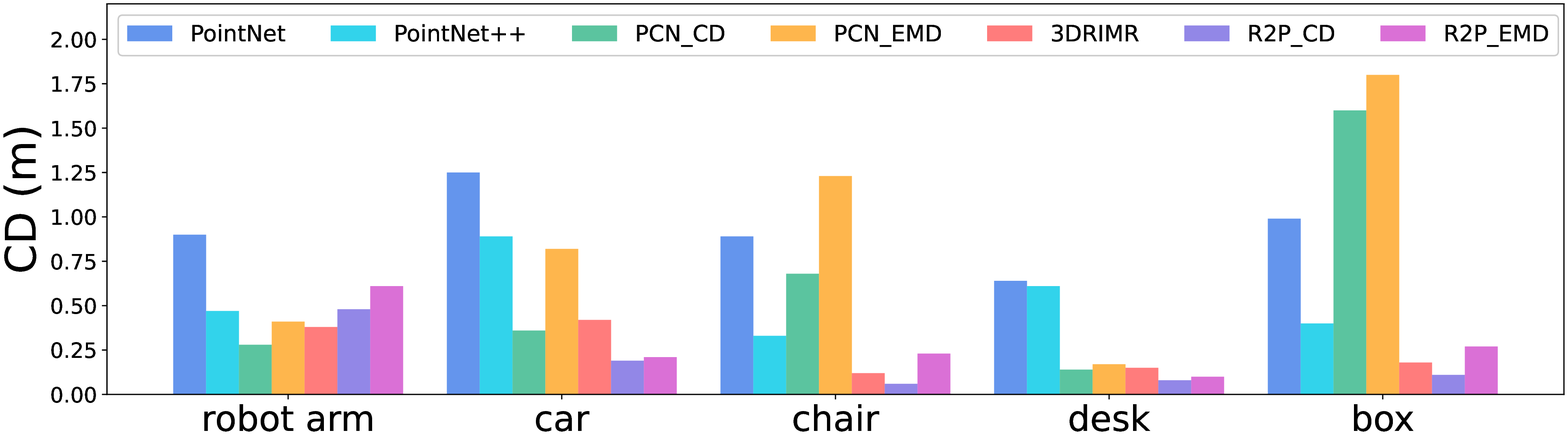}
				\setlength{\epsfxsize}{0.95\linewidth}
				\epsffile{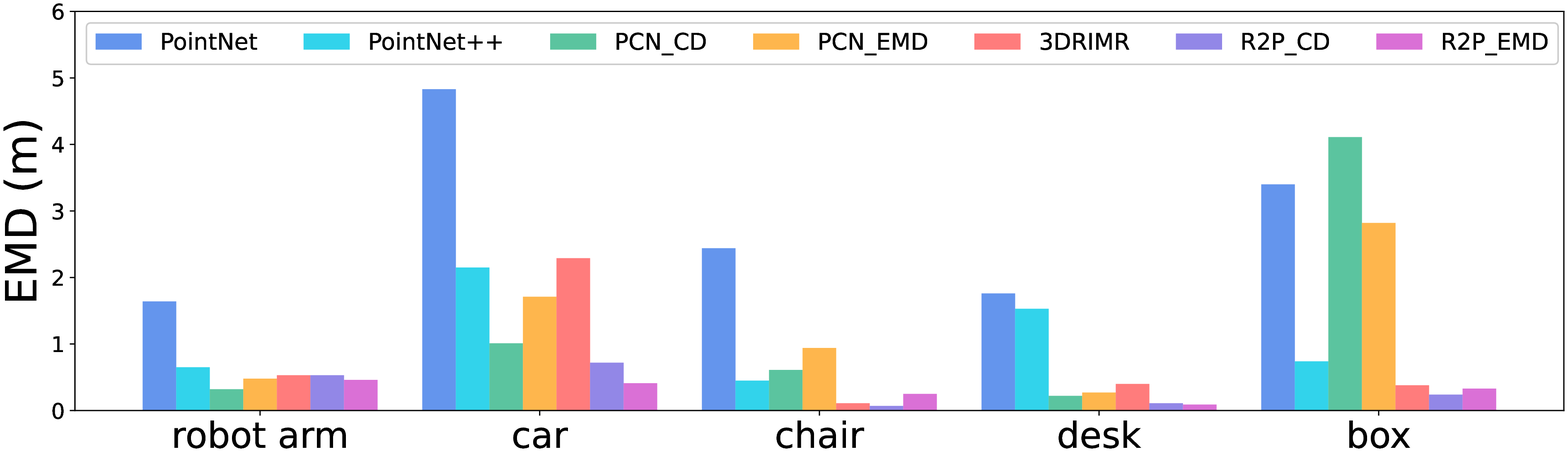}\\
				{}
			\end{center}
			\caption{Quantitative comparison on datasets of $5$ different objects using baseline methods and our method. Top: Chamfer Distance of the output point cloud and the ground truth. 
				Bottom: Earth Mover's Distance of the output point cloud and the ground truth.} \label{bar_chart}
			\label{6methods_plot}
		\end{minipage}
	}
\end{figure*}
%
We compare the performance of the two variants of our proposed R2P, 
labeled as R2P\_CD (both $d_1$ and $d_2$ use CD) and R2P\_EMD  (both $d_1$ and $d_2$ use EMD),
with the five baseline methods mentioned above in terms of CD and EMD.
The results are shown in Fig. \ref{6methods_plot}. 
We can see that except the case of robot arms, 
our methods always have the smallest CD or EMD loss among all the methods.
Even for the robot arms, the performance of our methods are similar to the other five methods. 

We further compare the results of all six methods by visually examining their output points clouds, and some of them are shown in Fig. \ref{pc_plots}. 
We can see that all the output point clouds are coarser than the ground truth point clouds 
and some of them lose many detailed shape characteristics. However,
R2P\_EMD can give the best shape reconstruction among all the methods, keeping most geometry characteristics.
Especially for some small objects with fine shape details like chairs,
only our method R2P\_EMD can reconstruct an object with accurate shape.

\begin{table}[h]
	\small
	\centering
	\begin{tabular}{l c c c c c}
		Input
		&
		\begin{minipage}[b]{0.16\columnwidth}
			\centering
			\raisebox{-.5\height}{\includegraphics[width=\linewidth]{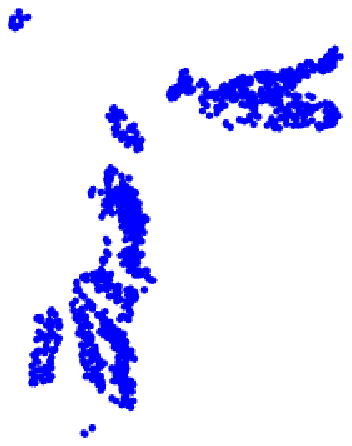}}
		\end{minipage}
		&
		\begin{minipage}[b]{0.16\columnwidth}
			\centering
			\raisebox{-.5\height}{\includegraphics[width=\linewidth]{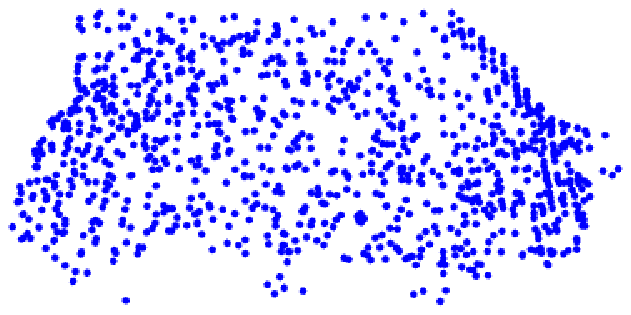}}
		\end{minipage}
		&
		\begin{minipage}[b]{0.16\columnwidth}
			\centering
			\raisebox{-.5\height}{\includegraphics[width=\linewidth]{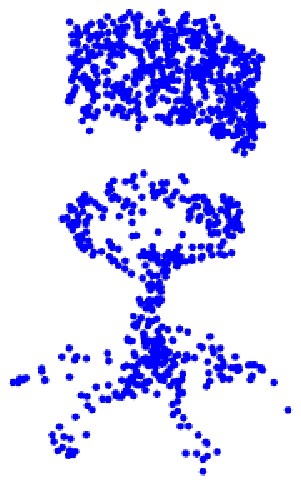}}
		\end{minipage}
		&
		\begin{minipage}[b]{0.16\columnwidth}
			\centering
			\raisebox{-.5\height}{\includegraphics[width=\linewidth]{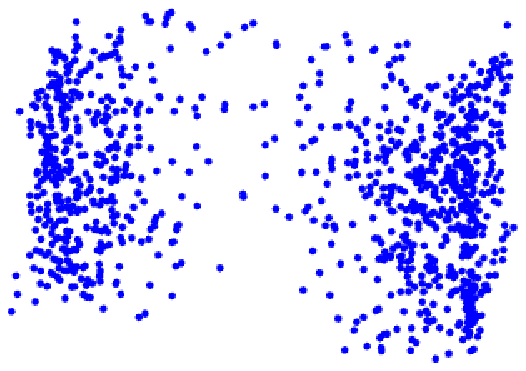}}
		\end{minipage}
		&
		\begin{minipage}[b]{0.16\columnwidth}
			\centering
			\raisebox{-.5\height}{\includegraphics[width=\linewidth]{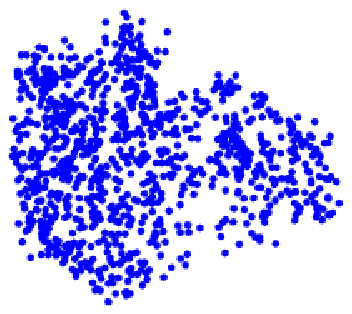}}
		\end{minipage}
		\\
		PointNet++
		&
		\begin{minipage}[b]{0.16\columnwidth}
			\centering
			\raisebox{-.5\height}{\includegraphics[width=\linewidth]{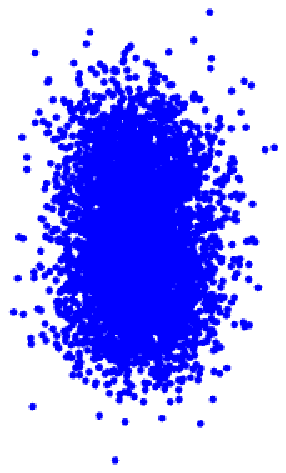}}
		\end{minipage}
		&
		\begin{minipage}[b]{0.16\columnwidth}
			\centering
			\raisebox{-.5\height}{\includegraphics[width=\linewidth]{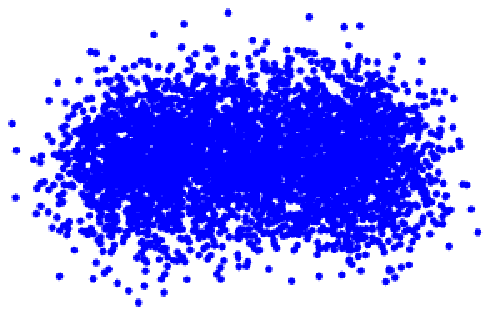}}
		\end{minipage}
		&
		\begin{minipage}[b]{0.16\columnwidth}
			\centering
			\raisebox{-.5\height}{\includegraphics[width=\linewidth]{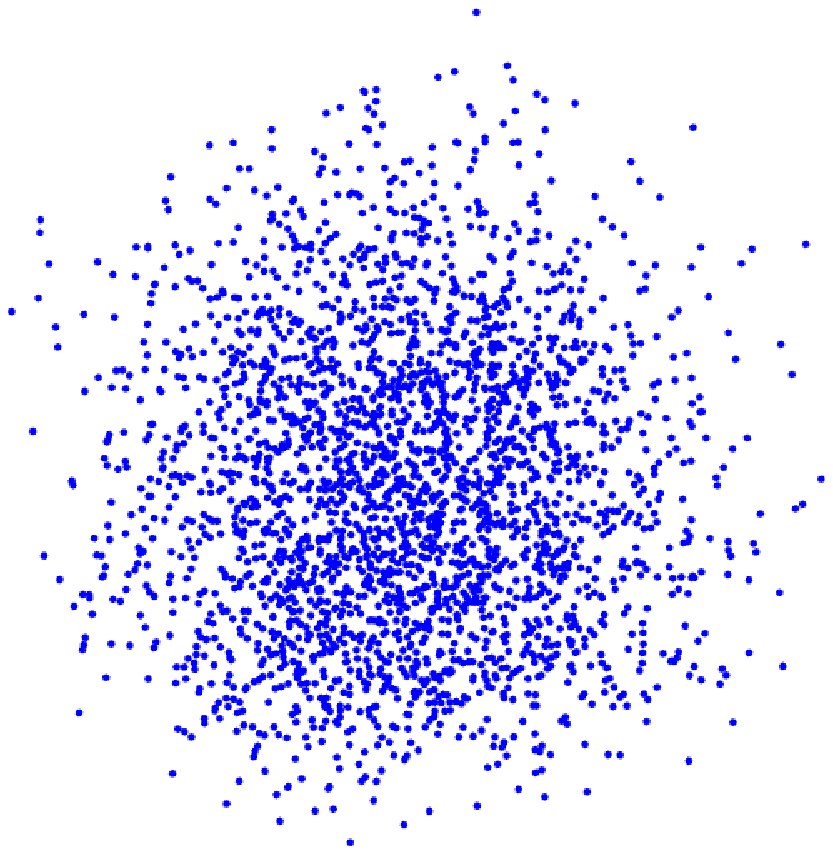}}
		\end{minipage}
		&
		\begin{minipage}[b]{0.16\columnwidth}
			\centering
			\raisebox{-.5\height}{\includegraphics[width=\linewidth]{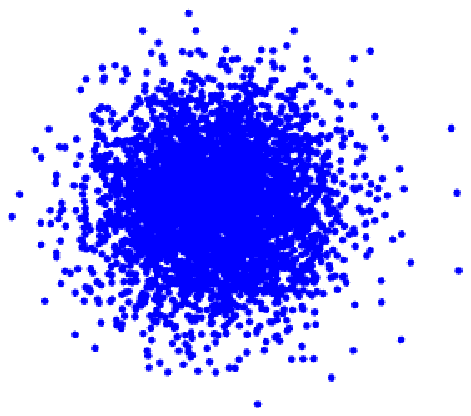}}
		\end{minipage}
		&
		\begin{minipage}[b]{0.16\columnwidth}
			\centering
			\raisebox{-.5\height}{\includegraphics[width=\linewidth]{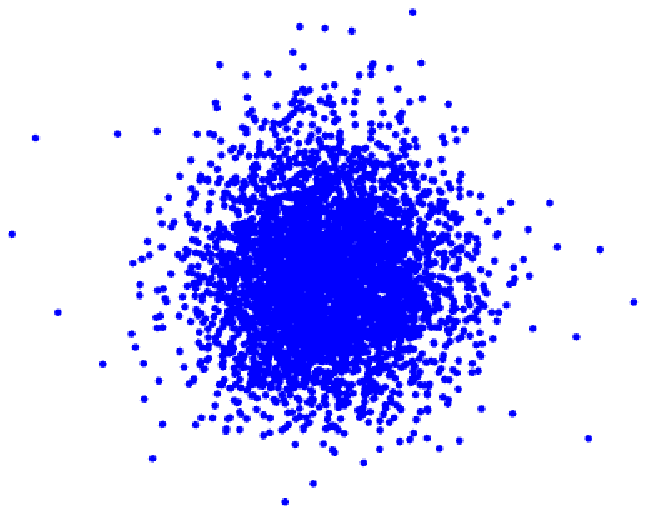}}
		\end{minipage}
	 	\\
	 	PCN\_EMD
	 	&
	 	\begin{minipage}[b]{0.16\columnwidth}
	 		\centering
	 		\raisebox{-.5\height}{\includegraphics[width=\linewidth]{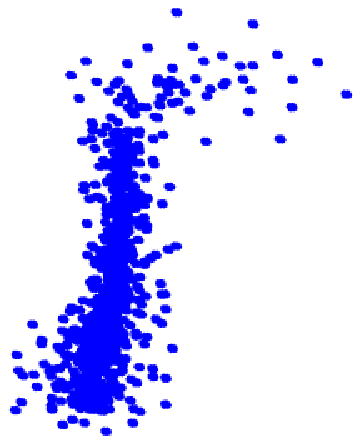}}
	 	\end{minipage}
	 	&
	 	\begin{minipage}[b]{0.16\columnwidth}
	 		\centering
	 		\raisebox{-.5\height}{\includegraphics[width=\linewidth]{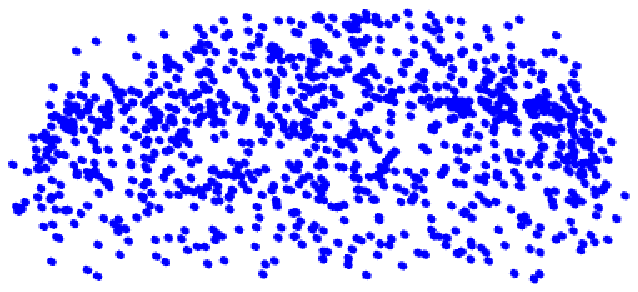}}
	 	\end{minipage}
	 	&
	 	\begin{minipage}[b]{0.16\columnwidth}
	 		\centering
	 		\raisebox{-.5\height}{\includegraphics[width=\linewidth]{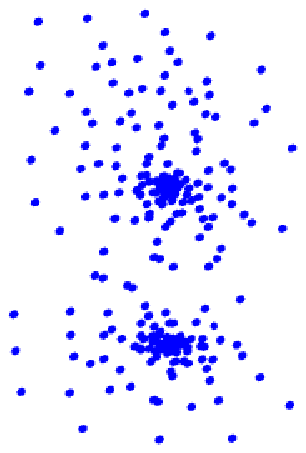}}
	 	\end{minipage}
	 	&
	 	\begin{minipage}[b]{0.16\columnwidth}
	 		\centering
	 		\raisebox{-.5\height}{\includegraphics[width=\linewidth]{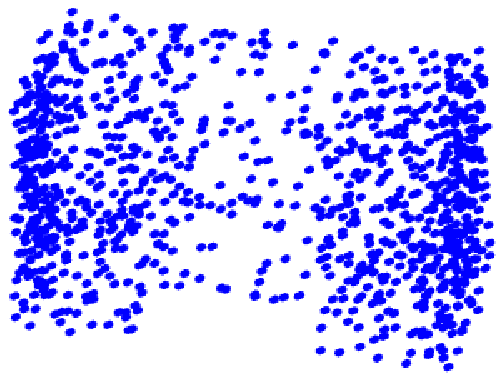}}
	 	\end{minipage}
	 	&
	 	\begin{minipage}[b]{0.16\columnwidth}
	 		\centering
	 		\raisebox{-.5\height}{\includegraphics[width=\linewidth]{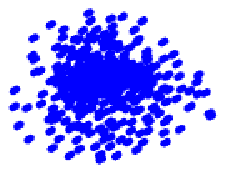}}
	 	\end{minipage}
 		\\
 		3DRIMR
 		&
 		\begin{minipage}[b]{0.16\columnwidth}
 			\centering
 			\raisebox{-.5\height}{\includegraphics[width=\linewidth]{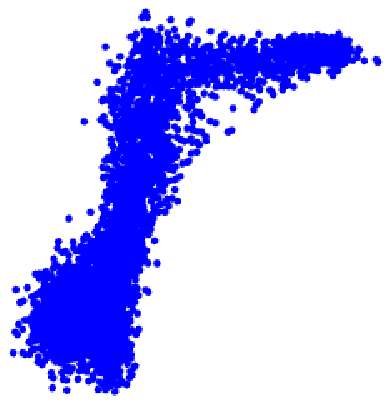}}
 		\end{minipage}
 		&
 		\begin{minipage}[b]{0.16\columnwidth}
 			\centering
 			\raisebox{-.5\height}{\includegraphics[width=\linewidth]{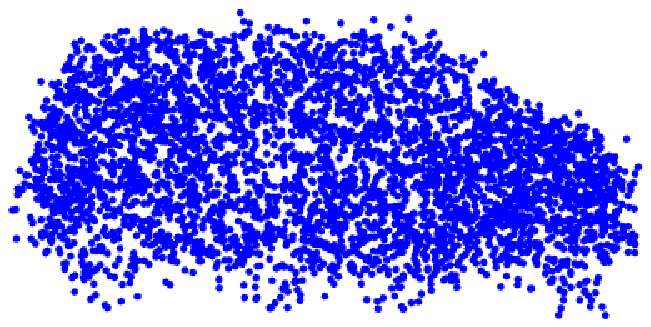}}
 		\end{minipage}
 		&
 		\begin{minipage}[b]{0.16\columnwidth}
 			\centering
 			\raisebox{-.5\height}{\includegraphics[width=\linewidth]{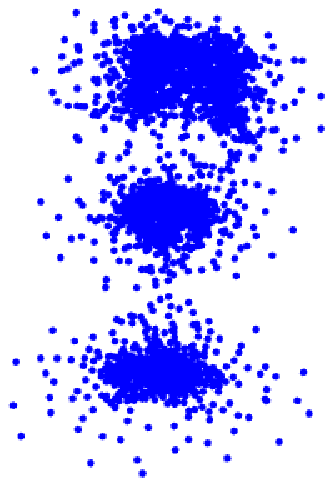}}
 		\end{minipage}
 		&
 		\begin{minipage}[b]{0.16\columnwidth}
 			\centering
 			\raisebox{-.5\height}{\includegraphics[width=\linewidth]{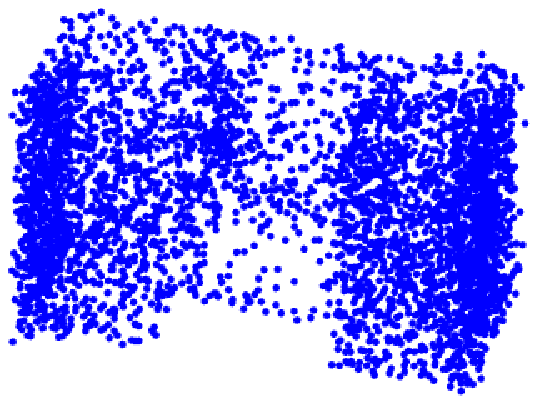}}
 		\end{minipage}
 		&
 		\begin{minipage}[b]{0.16\columnwidth}
 			\centering
 			\raisebox{-.5\height}{\includegraphics[width=\linewidth]{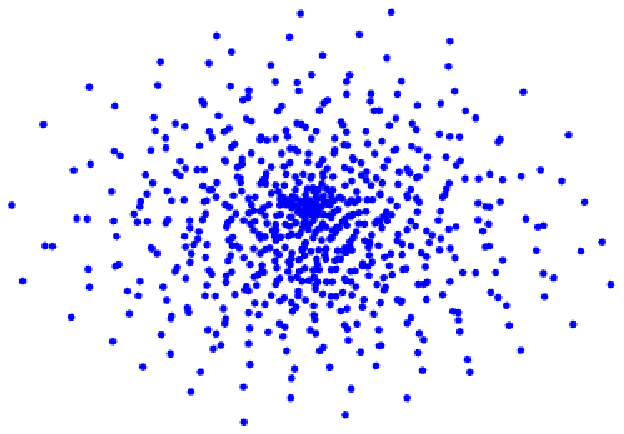}}
 		\end{minipage}
 		\\
 		R2P\_EMD
 		&
 		\begin{minipage}[b]{0.16\columnwidth}
 			\centering
 			\raisebox{-.5\height}{\includegraphics[width=\linewidth]{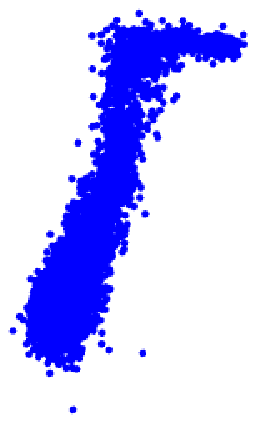}}
 		\end{minipage}
 		&
 		\begin{minipage}[b]{0.16\columnwidth}
 			\centering
 			\raisebox{-.5\height}{\includegraphics[width=\linewidth]{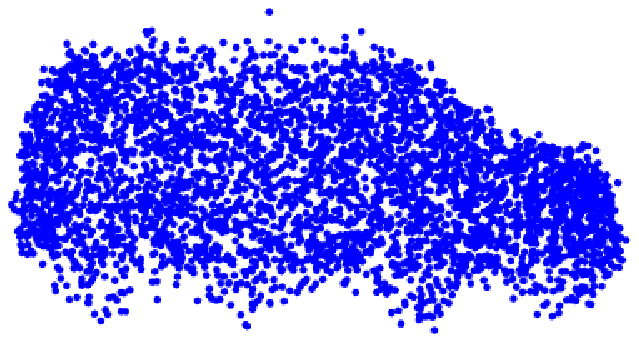}}
 		\end{minipage}
 		&
 		\begin{minipage}[b]{0.16\columnwidth}
 			\centering
 			\raisebox{-.5\height}{\includegraphics[width=\linewidth]{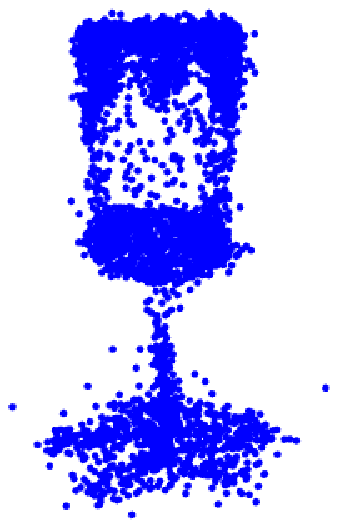}}
 		\end{minipage}
 		&
 		\begin{minipage}[b]{0.16\columnwidth}
 			\centering
 			\raisebox{-.5\height}{\includegraphics[width=\linewidth]{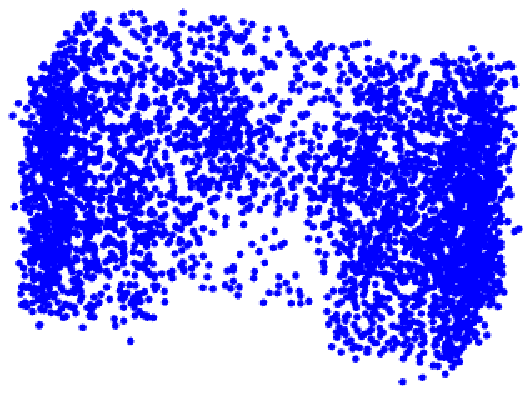}}
 		\end{minipage}
 		&
 		\begin{minipage}[b]{0.16\columnwidth}
 			\centering
 			\raisebox{-.5\height}{\includegraphics[width=\linewidth]{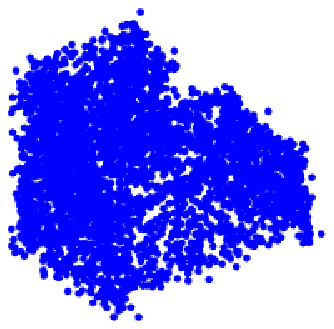}}
 		\end{minipage}
 		\\
 		Ground Truth
 		&
 		\begin{minipage}[b]{0.16\columnwidth}
 			\centering
 			\raisebox{-.5\height}{\includegraphics[width=\linewidth]{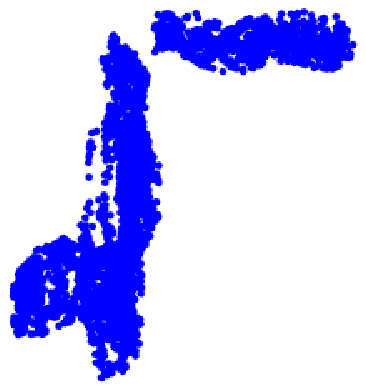}}
 		\end{minipage}
 		&
 		\begin{minipage}[b]{0.16\columnwidth}
 			\centering
 			\raisebox{-.5\height}{\includegraphics[width=\linewidth]{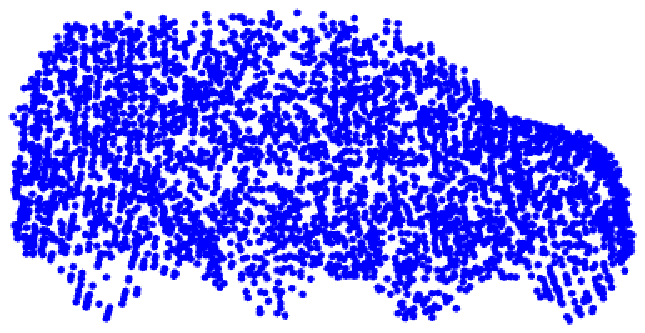}}
 		\end{minipage}
 		&
 		\begin{minipage}[b]{0.16\columnwidth}
 			\centering
 			\raisebox{-.5\height}{\includegraphics[width=\linewidth]{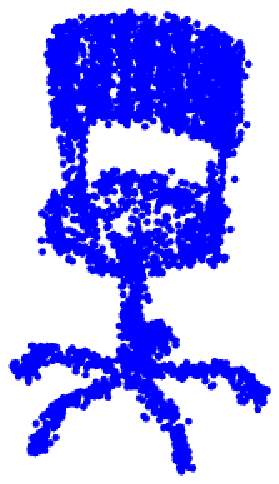}}
 		\end{minipage}
 		&
 		\begin{minipage}[b]{0.16\columnwidth}
 			\centering
 			\raisebox{-.5\height}{\includegraphics[width=\linewidth]{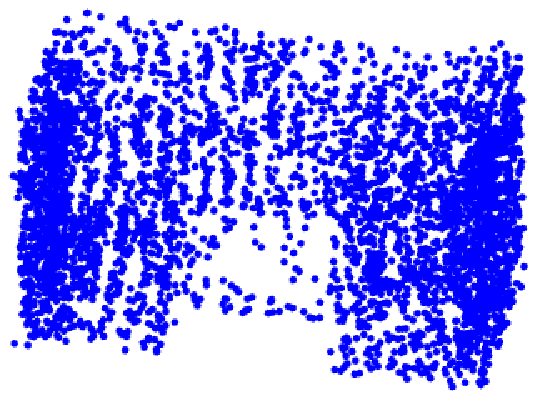}}
 		\end{minipage}
 		&
 		\begin{minipage}[b]{0.16\columnwidth}
 			\centering
 			\raisebox{-.5\height}{\includegraphics[width=\linewidth]{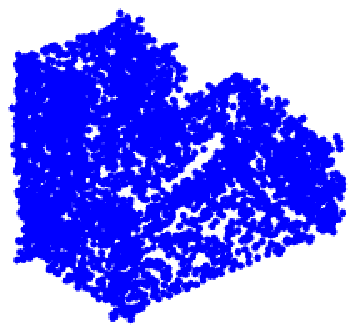}}
 		\end{minipage}
	
	\end{tabular}
	\captionof{figure}{Comparison of generated point clouds of different objects using different methods. 
		The output point clouds of PointNet, PCN\_CD, R2P\_CD are very similar to those of 
		PointNet++, PCN\_EMD, R2P\_EMD, respectively.  They are not shown here due to space limitations. }
\label{pc_plots}
\end{table}

\subsection{Performance of Different Loss Functions}

Loss function plays a quite important role in model training. 
A well-designed loss function can not only speed up the training process, 
but also affect the performance of a deep learning model.
On the other hand, a bad loss function may not converge even with a well-designed network model.
Hence, we need to carefully design our loss function used when training our model.
However, existing popular evaluation metrics to compare point clouds are CD and EMD,
which can only evaluate the overall distance between a pair of point clouds
but cannot effectively capture the similarity of their shapes. 

\begin{figure*}[htb!]
	\centerline{
		\begin{minipage}{0.99\linewidth}
			\begin{center}
				\setlength{\epsfxsize}{0.49\linewidth}
				\epsffile{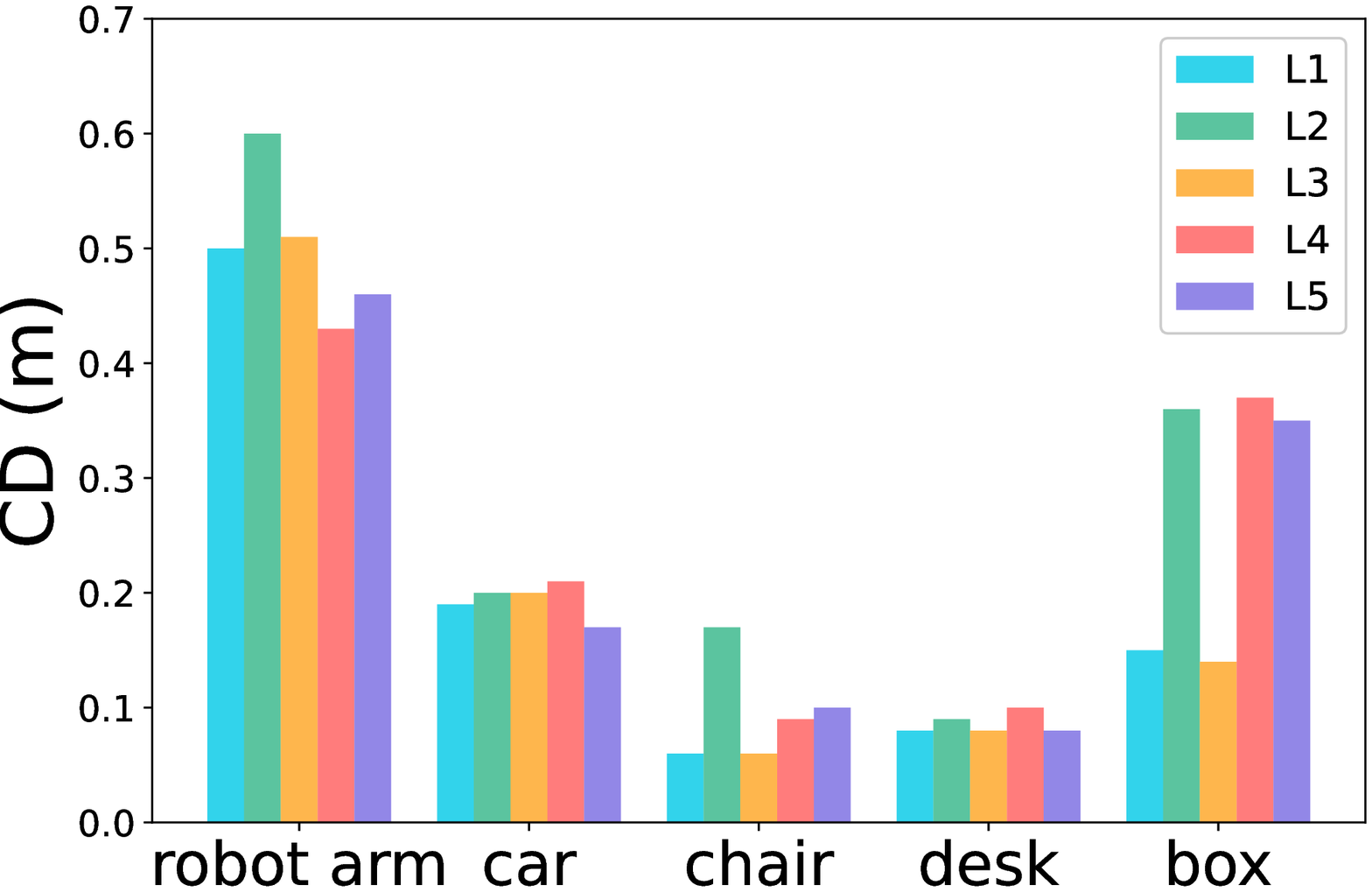}
				\setlength{\epsfxsize}{0.49\linewidth}
				\epsffile{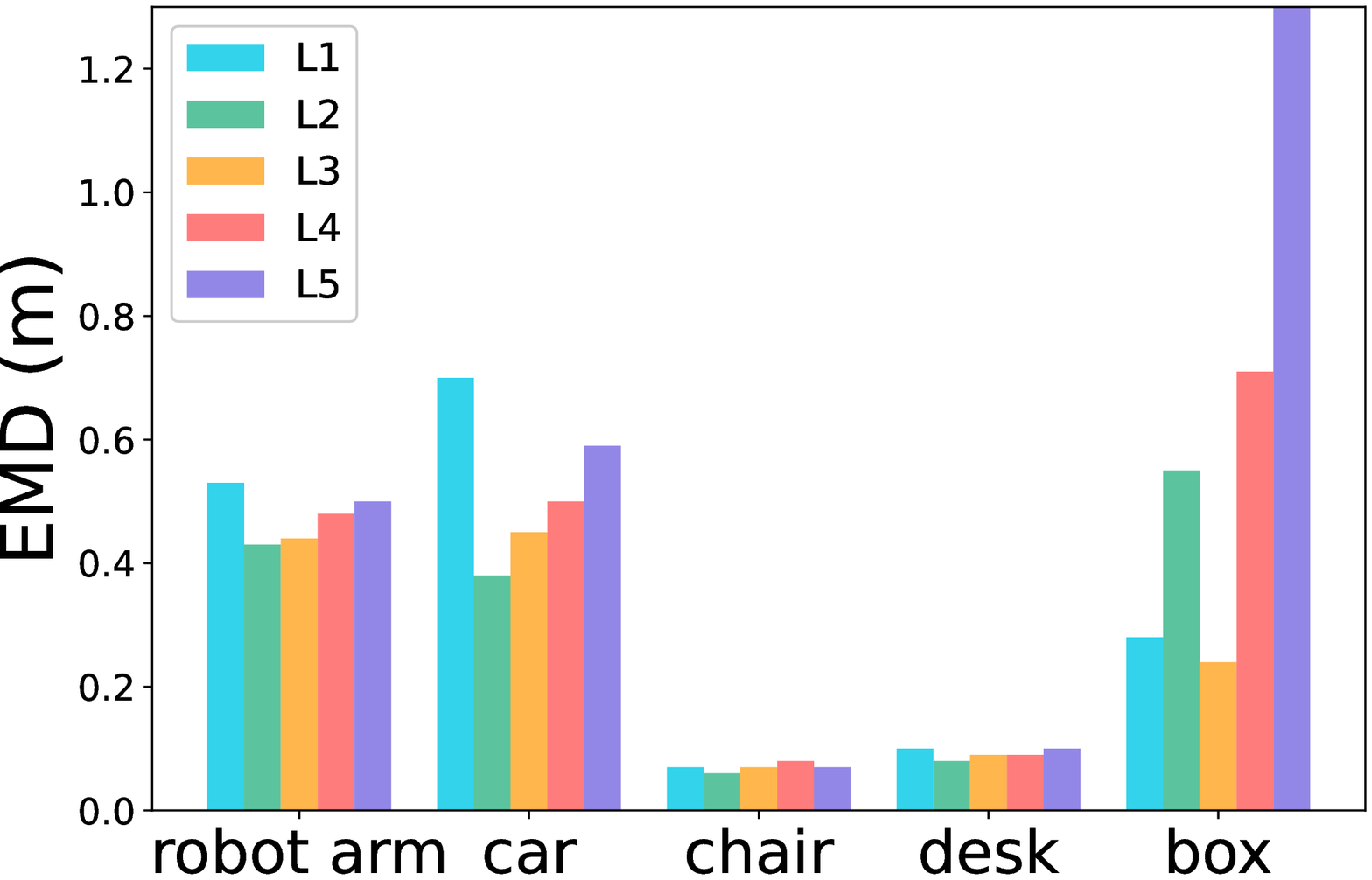}\\
				{}
			\end{center}
			\caption{Comparison of Different Loss Functions. 
				L1: both $d_1$ and $d_2$ are CD; L2: both $d_1$ and $d_2$ are EMD; L3: both $d_1$ and $d_2$ are CD+EMD; 
				L4: $d_1$ is CD and $d_2$ is EMD; L5: $d_1$ is EMD and $d_2$ is CD.}
			\label{loss_plot}
		\end{minipage}
	}
\end{figure*}

To search for a good loss function for R2P, 
we conduct a series of experiments with different combinations of these two metrics.
Except the loss functions used in training are different,
all other settings are all the same for these experiments. 
The quantitative results are shown in Fig. \ref{loss_plot}.
We can see that if we only use EMD in the loss function, which is L2 in the figure, 
the CD between output and ground truth point clouds will be large;
and except in the box experiment, 
if we only use CD in the loss function, which is L1 in the figure, 
the EMD between output and ground truth point clouds will also be large.
Hence, it makes sense to use the combination of CD and EMD, which is L3 in the figure,
which gives both small CD and EMD. 
However, since calculating EMD is very expensive and time-consuming,  
using CD to evaluate intermediate output $P_m$ and EMD to evaluate final output $P_o$, which is L4 in the figure,
is also a good choice for the sake of efficiency.


\noindent \textbf{Remarks}.
(1) CD and EMD are important metrics to evaluate the difference between two point clouds.
Generally speaking, 
small CD/EMD value means better reconstruction performance, and  vice versa.
However, 
due to the irregular format and lack-of-order information of point clouds, 
these two metrics are quite limited in terms of accurately indicating the shape difference between two point clouds,
and hence cannot accurately describe the performance of a point cloud reconstruction method.
Sometimes, a reconstructed point cloud may have larger CD or EMD though its shape is more similar to the ground truth point cloud.
For example, as we can see in Fig. \ref{6methods_plot}, the EMD of chair using 3DRIMR is smaller than R2P\_EMD, 
but from Fig. \ref{pc_plots}, we can see that the output of R2P\_EMD has more accurate shape of a chair.
Hence, we should not focus only on the values of CD or EMD when evaluating a method's reconstruction performance.
(2) We have also conducted a series of experiments to explore different designs of our network architecture, 
e.g., using different pooling methods to extract global features, applying discriminators during the training process, 
and deeper network architecture with more layers.
Nevertheless we find that the architecture shown in Fig. \ref{fig_G} performs the best and is efficient.

\section{CONCLUSIONS AND FUTURE WORK}\label{sec_conclusion}
We have proposed R2P, a deep learning model that generates 3D objects 
in the form of smooth, dense, and highly accurate point clouds with fine geometry details.
The inputs to R2P are directly 
converted from the 2D depth images that are generated from raw mmWave radar sensor data, and thus characterized by  
mutual inconsistency or errors in terms of orientation and shape.
We have demonstrated with extensive experiments  
that R2P significantly outperforms existing methods such as PointNet/PointNet++, PCN, and 3DRIMR. 
In addition, we have shown the importance of loss function design 
in the training of models for reconstructing point clouds. 
For future work, we will further improve our design and test it with large scale experiments in more practical environments 
and with more object categories. 

%
%
%
 \bibliographystyle{splncs04}
 \bibliography{references}
%

%
%
%
%

\end{document}